\documentclass[3p,times]{elsarticle}

\usepackage{ecrc}

\usepackage{amsmath}
\usepackage{amssymb}
\usepackage{amsfonts}
\usepackage{bm}
\usepackage[subrefformat=parens]{subcaption}

\firstpage{1}

\runauth{Yuta Kawakami et al.}

\jid{is}

\usepackage{amssymb}

\usepackage[figuresright]{rotating}

\begin{document}

\begin{frontmatter}



\dochead{}

\title{New Solutions Based on the Generalized Eigenvalue Problem for the Data Collaboration Analysis}


\author{Yuta Kawakami\fnref{label1}}
\author{Yuichi Takano\fnref{label2}}
\author{Akira Imakura\fnref{label2}}

\address[label1]{Degree Programs in Systems and Information Engineering, University of Tsukuba}
\address[label2]{Institute of Systems and Information Engineering, University of  Tsukuba}

\begin{abstract}
In recent years, the accumulation of data across various institutions has garnered attention for the technology of confidential data analysis, which improves analytical accuracy by sharing data between multiple institutions while protecting sensitive information. Among these methods, Data Collaboration Analysis (DCA) is noted for its efficiency in terms of computational cost and communication load, facilitating data sharing and analysis across different institutions while safeguarding confidential information. However, existing optimization problems for determining the necessary collaborative functions have faced challenges, such as the optimal solution for the collaborative representation often being a zero matrix and the difficulty in understanding the process of deriving solutions. This research addresses these issues by formulating the optimization problem through the segmentation of matrices into column vectors and proposing a solution method based on the generalized eigenvalue problem. Additionally, we demonstrate methods for constructing collaborative functions more effectively through weighting and the selection of efficient algorithms suited to specific situations. Experiments using real-world datasets have shown that our proposed formulation and solution for the collaborative function optimization problem achieve superior predictive accuracy compared to existing methods.
\end{abstract}

\begin{keyword}
Confidential Data Analysis \sep Distributed Data \sep Data Collaboration \sep Mathematical Modeling \sep Machine Learning

\end{keyword}

\end{frontmatter}


\section{Introduction}
\label{sec:introduction}

\subsection{Background}
In recent years, the advancement of IT technologies has led to an increase in large-scale and distributed data. This development has enabled data analysis from multiple perspectives. However, to enhance the accuracy of analysis, a sufficient amount of data is necessary. In situations where data is sparse and distributed, there are two primary analysis methods: individual analysis, which involves analyzing the original data within each institution, and centralized analysis, where the original data from all institutions is collected and analyzed at one location.

Among these methods, the practical application of centralized analysis, which can increase data volume and thus improve analytical accuracy compared to individual analysis, is highly desirable. However, the sharing and aggregation of original data pose challenges in terms of protecting confidential information. To address this, Data Collaboration Analysis (DCA) was proposed by Imakura and Sakurai~\cite{dca-ori}. In DCA, the original data is abstracted at each institution using dimensionality reduction techniques that are computationally efficient. This abstracted data, which has been safeguarded to protect confidential information, is then aggregated from all institutions. The confidentiality of the original data in this abstracted form has been proven by Imakura et al.~\cite{dca-sec}, based on security against both internal and external attack scenarios. A notable point is that analysis using this aggregated abstracted data, without directly aggregating the original data, can improve analytical accuracy similar to centralized analysis.

In DCA, this abstracted data cannot be directly computed with abstracted data sent from other institutions. Therefore, a collaborative function is needed to transform the abstracted data from each institution so that it can be computed together. Existing DCA methods~\cite{dca-ori} formulated an optimization problem to find this collaborative function, replacing the problem with a minimal perturbation problem. However, the conventional optimization problem for finding the collaborative function faced challenges, such as the optimal solution for the collaborative representation often being a zero matrix and the difficulty in understanding the derivation process of the solution. This study proposes a formulation of this optimization problem as a constrained optimization problem by segmenting matrices into column vectors and presents a solution to these challenges.

\subsection{Related works}
\label{sec:related_works}

In this subsection, we overview existing methods related to confidential data analysis and DCA.

The primary goal of DCA is to resolve the issue of protecting confidential information in centralized analysis. Representative methods in confidential data analysis include secure computation, which secures confidential information based on cryptography \cite{sec-4, sec-5, sec-2, sec-3, sec-1}, differential privacy, which adds noise to prevent the leakage of training data \cite{dif-2, dif-1, dif-3}, and federated learning, which involves learning models locally and only sharing model update information \cite{fed-2, fed-1}. However, each of these methods has its own challenges: secure computation is computationally expensive, differential privacy can degrade data quality, and federated learning faces issues with communication burden and the complexity of data sharing. In contrast, DCA reduces the costs of computation, communication, and data sharing, without degrading data quality due to noise, hence holding a superior position in confidential data analysis methods.

DCA has shown high performance in specific scenarios, such as anomaly detection using kernel functions \cite{dca-novelty} and evaluation of specific data distribution situations \cite{dca-dist}. These studies demonstrate that DCA can provide superior analytical results over individual institutions, even in situations where data is distributed across different organizations.

Generally, in DCA, a dataset constructed randomly based on standard normal distribution, known as anchor data, is used. The creation of anchor data using over-sampling methods like SMOTE \cite{dca-smote} has been shown to generate data close to the distribution of the original dataset, thereby enhancing the accuracy and interpretability of the analysis.

Research on the optimization problem of collaborative functions in DCA has been conducted in the context of optimization on manifolds \cite{dca-man}. This method offers higher analytical accuracy than the existing collaborative function optimization method using minimal perturbation problems \cite{dca-ori}, but it is limited to small-scale problems and has the challenge of long computation times. This paper proposes a new method that, while using a similar formulation to existing models \cite{dca-ori, dca-man}, achieves faster and more accurate analysis for a wider range of problems through mathematical transformations and innovative solution methods.

Furthermore, DCA has been proposed in different frameworks, such as methods combining Node2Vec \cite{dca-n2v} and non-readable DCA \cite{dca-nri}. These methods aim to improve analytical accuracy and security, but they face challenges such as computational costs and the complexity of processing required by each institution. This research overcomes these challenges, offering a more efficient and practical analytical framework.

This paper builds upon the original DCA framework \cite{dca-ori} and proposes a new formulation and solution method for the collaborative function optimization problem. This approach is not constrained by the dimensions of the data or the number of institutions and is applicable to problems ranging from small to large scale.

\subsection{Our contribution}
In the proposed method, under the premise of clear solution, a different approach from existing optimization methods is adopted to determine the collaborative function. By exploiting the characteristic of finding the collaborative function for each column vector, the proposed method introduces techniques that achieve higher precision and efficient solutions in analysis, thereby enhancing performance in DCA.

The main contributions of this study are as follows:
\begin{enumerate}
	\item Simplification of the solution process for the collaborative function optimization problem.
	\item In numerical experiments using real-world image and tabular datasets, the proposed formulation and solution method for the collaborative function achieve higher performance than existing methods.
	\item Efficient determination of the collaborative function by selecting solutions according to the situation.
\end{enumerate}

Next, the structure of this paper is described. The paper is composed of seven chapters. Chapter 2 introduces notation for various mathematical symbols and terms used throughout the paper. Chapter 3 introduces existing methods for the optimization of collaborative functions in DCA. Chapter 4 introduces the proposed method for the optimization of collaborative functions in DCA. Chapter 5 explains the setup of numerical experiments using real-world datasets. Chapter 6 presents the results and discussion of the numerical experiments. Finally, Chapter 7 concludes the paper and discusses future challenges.

\section{Notation}
We define the set of integers from 1 to $n'$ as follows:
\begin{equation}
	[n'] \coloneqq \{1,2,\ldots,n'\} \notag
\end{equation}

For a row vector $\bm{x}'_d \in \mathbb{R}^{1 \times p}~(d \in [n'])$, we define the matrix composed by stacking these vectors vertically as:
\begin{equation}
	\left(
		\bm{x}'_d
	\right)_{d \in [n']}
	\coloneqq 
	\begin{pmatrix}
		\bm{x}'_1 \\
		\bm{x}'_2 \\
		\vdots \\
		\bm{x}'_{n'}
	\end{pmatrix} \in \mathbb{R}^{n' \times p} \notag
\end{equation}

\section{Existing methods}
\label{sec:e-methods}

In this section, we introduce the existing methods for the optimization of collaborative functions in DCA \cite{dca-ori}.

Let $m$ represent the dimension number of each data row in the original data, $N$ the total number of institutions, ${n_i}$ the number of original data held by institution $i$, and $n$ the total number of original data held by all institutions (where ${n=\sum_{i=1}^{N}n_i}$). In addition, let $\bm{X}_i$ denote the original data of institution $i$, and $\bm{L}_i$ represent the corresponding teacher labels in supervised learning. $\bm{X}_i$ and $\bm{L}_i$ are structured as following matrices. Furthermore, $m_\mathrm{label}$ indicates the dimension number of the teacher labels.

\begin{equation}
  \bm{X}_i = \left(
		\bm{x}_{id} 
	\right)_{d \in [n_i]} \in \mathbb{R}^{{n_i} \times {m}}, 
  \bm{L}_i = \left(
    \bm{l}_{id}
	\right)_{d \in [n_i]} \in \mathbb{R}^{{n_i} \times {m_{\mathrm{label}}}}
\end{equation}

Additionally, in DCA, a common dataset known as anchor data, $\bm{X}^\mathrm{anc} \in \mathbb{R}^{{r} \times {{m}}}$, is shared among all institutions. The anchor data is constructed either from open data or as random dummy data. In DCA, the original data ${\bm{X}_i}$ of each institution is not aggregated directly. Instead, an intermediate representation ${\tilde{\bm{X}_i}}$, ${\tilde{\bm{X}}^\mathrm{anc}_i}$, transformed by a unique function specific to each institution, is aggregated at a single analysis institution. Moreover, if the function used for this transformation is denoted as the abstraction function $f_i$, then the intermediate representation is expressed as follows. Here, $\tilde{m}_i$ represents the dimension number of the data in the intermediate representation.
\begin{equation}
  \tilde{\bm{X}}_i = \left(
    f_i(\bm{x}_{id})
	\right)_{d \in [n_i]} \in \mathbb{R}^{{n_i} \times {\tilde{m}_i}}, \tilde{\bm{X}}^{\mathrm{anc}}_i = \left(
    f_i({\bm{x}^\mathrm{anc}_{d}})
	\right)_{d \in [r]} \in \mathbb{R}^{{r} \times {\tilde{m}_i}}
\end{equation}

At the analysis institution where the intermediate representations are aggregated, a collaborative function $g_i$ is constructed such that $g_i(f_i({\bm{x}^\mathrm{anc}_{d}})) \approx g_{i'}(f_{i'}({\bm{x}^\mathrm{anc}_{d}}))(i \neq i')$. Here, $\hat{m}$ represents the dimension number of the data after transforming the intermediate representation. To construct the collaborative function, we use a matrix $\bm{G}_i \in \tilde{m}_i \times \hat{m}$ where $g_i(f_i({\bm{x}^\mathrm{anc}_{d}})) = f_i({\bm{x}^\mathrm{anc}_{d}})G_i$, and employ a solution method for the minimal perturbation problem based on a rank $\hat{m}_i$ approximate singular value decomposition algorithm. Note that $\bm{U}_{\hat{m}} \in \mathbb{R}^{r \times \hat{m}}$.  This solution method is used for the optimization problem of the matrix.

\begin{eqnarray}
	\left(
		\tilde{\bm{X}}^\mathrm{anc}_1,\tilde{\bm{X}}^\mathrm{anc}_2,\cdots,\tilde{\bm{X}}^\mathrm{anc}_N
	\right) \approx \bm{U}_{\hat{m}}\bm{\Sigma}_{\hat{m}}\bm{V}_{\hat{m}}^\top \\
	\bm{G}_i = (\tilde{\bm{X}}^\mathrm{anc}_i)^\dagger \bm{U}_{\hat{m}}
\end{eqnarray}

At the analysis institution, the intermediate representation is transformed into the following collaborative representation $\hat{\bm{X}}_i$. Analysis is then performed using this collaborative representation.
\begin{equation}
  \hat{\bm{X}}_i = \left(
    g_i(f_i({\bm{x}_{id}}))
	\right)_{d \in [n_i]} = \left(
    f_i({\bm{x}_{id}}) \bm{G}_i
	\right)_{d \in [n_i]} = \tilde{\bm{X}}_i \bm{G}_i \in \mathbb{R}^{{n_i} \times {\hat{m}}}
\end{equation}

\section{Proposed methods}
\label{sec:methods}

In this section, we introduce the proposed method for the optimization of collaborative functions in DCA.

The proposed method employs a generalized eigenvalue problem. First, the matrix $\bm{G}_i$, which corresponds to the collaborative function, is decomposed into column vectors as follows:
\begin{equation}
	\label{G2vector}
	\bm{G}_i = \left( \bm{g}_{i1} \hdots \bm{g}_{ij} \hdots \bm{g}_{i\hat{m}} \right)
\end{equation}
By decomposing the matrix into column vectors in this way, for each $j \in [\hat{m}]$, it is possible to replace the minimization problem of the matrix residual in the existing method \cite{dca-ori} with the minimization problem of the residual of the column vectors.

First, matrices $\bm{A}_{\tilde{X}}$ and $\bm{B}_{\tilde{X}}$ are defined as follows:
\begin{eqnarray}
	\bm{A}_{\tilde{X}} = \left(
		\begin{array}{cccc}
			2(N-1)\tilde{\bm{X}}_{1}^\mathrm{anc\top}\tilde{\bm{X}}_{1}^\mathrm{anc} & -2\tilde{\bm{X}}_{1}^\mathrm{anc\top}\tilde{\bm{X}}_{2}^\mathrm{anc} & \hdots & -2\tilde{\bm{X}}_{1}^\mathrm{anc\top}\tilde{\bm{X}}_{N}^\mathrm{anc} \\ 
			-2\tilde{\bm{X}}_{2}^\mathrm{anc\top}\tilde{\bm{X}}_{1}^\mathrm{anc} & 2(N-1)\tilde{\bm{X}}_{2}^\mathrm{anc\top}\tilde{\bm{X}}_{2}^\mathrm{anc} & & \vdots \\ 
			\vdots & & \ddots & \\ 
			-2\tilde{\bm{X}}_{N}^\mathrm{anc\top}\tilde{\bm{X}}_{1}^\mathrm{anc} & \hdots & & 2(N-1)\tilde{\bm{X}}_{N}^\mathrm{anc\top}\tilde{\bm{X}}_{N}^\mathrm{anc}
		\end{array}
	\right) \\
	\label{eig-B} \bm{B}_{\tilde{X}} = \left(
		\begin{array}{cccc}
			\tilde{\bm{X}}_{1}^\mathrm{anc\top}\tilde{\bm{X}}_{1}^\mathrm{anc} & & & O \\ 
			& \tilde{\bm{X}}_{2}^\mathrm{anc\top}\tilde{\bm{X}}_{2}^\mathrm{anc} & & \\ 
			& & \ddots & \\ 
			O & & & \tilde{\bm{X}}_{N}^\mathrm{anc\top}\tilde{\bm{X}}_{N}^\mathrm{anc}
		\end{array}
	\right)
\end{eqnarray}
Moreover, $\bm{v}_j$ is a vector constructed by vertically concatenating the $j$-th column of all institutions' column vectors $\bm{g}_{ij}$.
\begin{equation}
	\label{eig-v-g}
	\bm{v}_j = \left(
		\bm{g}_{ij}
	\right)_{i \in [N]}
\end{equation}
Let's denote any two institutions as $i$ and $i'$. For each $j \in [\hat{m}]$, we formulate the problem with the difference in all collaborative representations as the objective function, and the norm of the collaborative representation as the constraint:
\begin{equation}
	\label{eig-problem}
	\begin{split}
		\min_{\bm{g}_{ij}} F(\bm{v}_j) &= \sum^N_{i=1}\sum^N_{i'=1}\| \tilde{\bm{X}}_i^\mathrm{anc}\bm{g}_{ij} - \tilde{\bm{X}}_{i'}^\mathrm{anc}\bm{g}_{i'j} \|^2_2
		= \bm{v}_j^\top \bm{A}_{\tilde{X}} \bm{v}_j \\
		\mathrm{s.t.} \; C(\bm{v}_j) &= \sum^N_{i=1}\| \tilde{\bm{X}}_i^\mathrm{anc}\bm{g}_{ij} \|^2_2 - 1 
		= \bm{v}_j^\top \bm{B}_{\tilde{X}} \bm{v}_j - 1= 0
	\end{split}
\end{equation}

Using the method of Lagrange multipliers, we find that:
\begin{equation}
	\label{eig-l-3}
	\bm{A}_{\tilde{X}} \bm{v}_j = \lambda_j \bm{B}_{\tilde{X}} \bm{v}_j \quad (\bm{v}_j^\top \bm{B}_{\tilde{X}} \bm{v}_j = 1)
\end{equation}
This leads to a generalized eigenvalue problem. Therefore, with the generalized eigenvector $\bm{v}_j$ and the generalized eigenvalue $\lambda_j$, the objective function attains an extremum under the constraint.

Furthermore, considering $\bm{v}_j$ as the generalized eigenvector, when we expand the objective function $F(\bm{v}_j)$ using equation (\ref{eig-l-3}), the value of the objective function coincides with the generalized eigenvalue $\lambda_j$.
\begin{equation}
	\label{eig-target-lambda}
	F(\bm{v}_j) = \bm{v}_j^\top A_{\tilde{X}} \bm{v}_j = \lambda \bm{v}_j^\top B_{\tilde{X}} \bm{v}_j = \lambda_j
\end{equation}
Therefore, when arranging the extremum values of the objective function $F(\bm{v}_j)$ in ascending order, they correspond to the generalized eigenvalues $\lambda_j$ arranged in the same manner. As shown in equation (\ref{eig-v-g}), since $\bm{v}_j$ is formed by vertically concatenating the $j$-th column of all institutions' column vectors $\bm{g}_{ij}$, the collaborative function can be determined from $\bm{v}_j$ corresponding to the arranged $\lambda_j$. Furthermore, as $j \in [\hat{m}]$, the $\lambda_j$ are determined up to the $\hat{m}$-th smallest $(\lambda_1 < \lambda_2 < \hdots < \lambda{\hat{m}})$.

Finally, we summarize the procedure for determining the collaborative function using the formulation involving the generalized eigenvalue problem. In this method, we provided a constrained minimization problem to minimize the difference in the collaborative representations of each institution for the anchor data, under the norm constraint of the collaborative representations of the anchor data. 

The steps for determining the collaborative function are outlined as follows:

\begin{enumerate}
	\item Calculate matrices $\bm{A}_{\tilde{X}}$ and $\bm{B}_{\tilde{X}}$ through matrix computations.
	\item \label{eig-conc-enu} Solve the generalized eigenvalue problem for $\bm{A}_{\tilde{X}}$ and $\bm{B}_{\tilde{X}}$ as per equation (\ref{eig-l-3}), and determine the generalized eigenvalues $\lambda_j$ up to the $\hat{m}$-th smallest.
	\item From the $\bm{v}_j$ corresponding to the $\lambda_j$ obtained in step \ref{eig-conc-enu}, assign elements corresponding to each institution and construct the collaborative function.
\end{enumerate}

\subsection{Effective Construction Method for the Collaborative Function}
Previously, in the optimization problem of the collaborative function divided into column vectors, the construction of the collaborative function involved determining vectors for each column $j \in [\hat{m}]$, with the dimension number $\hat{m}$ predetermined. Here, we propose an effective construction method for the collaborative function using the generalized eigenvalue problem. As mentioned in equation (\ref{eig-target-lambda}), in the collaborative function optimization problem employing the generalized eigenvalue problem, the value of the objective function, representing the difference in column $j$ of all collaborative representations of the anchor data, is the generalized eigenvalue $\lambda_j$. This implies that the generalized eigenvalue $\lambda_j$ indicates the difference in all collaborative representations of the anchor data and the inaccuracy of the collaborative function. This understanding forms the premise for the effective construction method of the collaborative function, namely the weighting method, which we introduce next.

The column vector $\hat{\bm{x}}_{ij}$ of the collaborative representation $\hat{\bm{X}}_i$ represents the $j$-th feature of the collaborative representation, and $\hat{\bm{x}}_{ij}$ is generated by the column vector $\bm{g}_{ij}$. In this case, if the accuracy of $\bm{g}_{ij}$ is poor, it negatively impacts the analysis due to the corresponding feature $\hat{\bm{x}}_{ij}$. Therefore, in the weighting method, the following weighting function $w(\lambda_j)$ is used to attenuate the scale of features that have a significant negative impact before the analysis of the collaborative representation, as in $\hat{\bm{x}}_{ij} \leftarrow w(\lambda_j)\hat{\bm{x}}_{ij}$.

\begin{equation}
	w(\lambda_j) = \exp{\left( - \frac{\lambda_j - \lambda_1}{\lambda_{\hat{m}} - \lambda_1} \right)}
\end{equation}
The weighting method is effective in cases of model learning with regularization terms or other penalties on the weights. However, it is not meaningful for methods like random forests where scale is not a factor.

\subsection{Solution Reducing to Singular Value Decomposition}
Next, we introduce a solution method that employs QR decomposition and Singular Value Decomposition (SVD) to efficiently solve the collaborative function optimization problem using the generalized eigenvalue problem. First, QR decomposition is applied to $\tilde{\bm{X}}_i^{\mathrm{anc}}$ to replace the generalized eigenvalue problem with a special eigenvalue problem. When QR decomposition is applied to $\tilde{\bm{X}}_i^{\mathrm{anc}}$, it is decomposed into the product of two matrices as follows.
\begin{equation}
	\tilde{\bm{X}}_i^{\mathrm{anc}} = \bm{Q}_i^{\mathrm{anc}}\bm{R}_i^{\mathrm{anc}}
\end{equation}
In this case, $\bm{Q}_i^{\mathrm{anc}}$ is an orthogonal matrix such that $\bm{Q}_i^{\mathrm{anc}\top}\bm{Q}_i^{\mathrm{anc}}=\bm{I}$, where $\bm{I}$ is the identity matrix, and $\bm{R}_i^{\mathrm{anc}}$ is an upper triangular matrix. Following equation means the matrix sizes of $\tilde{\bm{X}}_i^\mathrm{anc}$ and $\bm{Q}_i^{\mathrm{anc}}$, and the vector sizes of $\bm{g}_{ij}$ and $(\bm{R}_i^{\mathrm{anc}}\bm{g}_{ij})$ to be equal. 
\begin{equation}
	\label{qr-anchor-col-rep}
	\tilde{\bm{X}}_i^\mathrm{anc}\bm{g}_{ij} = \bm{Q}_i^{\mathrm{anc}}(\bm{R}_i^{\mathrm{anc}}\bm{g}_{ij})
\end{equation}
Here, $\bm{W}_Q$, $\bm{A}_Q$, $\bm{B}_Q$, and ${\bm{v}_j}'$ are given by:
\begin{eqnarray}
	\bm{W}_Q = \left(
    \begin{array}{cccc}
      \bm{Q}_{1}^\mathrm{anc} & \bm{Q}_{2}^\mathrm{anc} & \hdots & \bm{Q}_{N}^\mathrm{anc}
    \end{array}
  \right) \\
	\bm{A}_Q = \left(
		\begin{array}{cccc}
			2(N-1)\bm{Q}_{1}^\mathrm{anc\top}\bm{Q}_{1}^\mathrm{anc} & -2\bm{Q}_{1}^\mathrm{anc\top}\bm{Q}_{2}^\mathrm{anc} & \hdots & -2\bm{Q}_{1}^\mathrm{anc\top}\bm{Q}_{N}^\mathrm{anc} \\ 
			-2\bm{Q}_{2}^\mathrm{anc\top}\bm{Q}_{1}^\mathrm{anc} & 2(N-1)\bm{Q}_{2}^\mathrm{anc\top}\bm{Q}_{2}^\mathrm{anc} & & \vdots \\ 
			\vdots & & \ddots & \\ 
			-2\bm{Q}_{N}^\mathrm{anc\top}\bm{Q}_{1}^\mathrm{anc} & \hdots & & 2(N-1)\bm{Q}_{N}^\mathrm{anc\top}\bm{Q}_{N}^\mathrm{anc}
		\end{array}
	\right) = -2(\bm{W}_{Q}^{\top}\bm{W}_{Q}) + 2N\bm{I} \\
	\label{eig-B-Q} \bm{B}_Q = \left(
		\begin{array}{cccc}
			\bm{Q}_{1}^\mathrm{anc\top}\bm{Q}_{1}^\mathrm{anc} & & & O \\ 
			& \bm{Q}_{2}^\mathrm{anc\top}\bm{Q}_{2}^\mathrm{anc} & & \\ 
			& & \ddots & \\ 
			O & & & \bm{Q}_{N}^\mathrm{anc\top}\bm{Q}_{N}^\mathrm{anc}
		\end{array}
	\right) = I \\
	\label{eig-v-g-prime}
	{\bm{v}_j}' = \left(
		\begin{array}{c}
			R_1^{\mathrm{anc}}\bm{g}_{1j} \\
			R_2^{\mathrm{anc}}\bm{g}_{2j} \\ 
			\vdots \\ 
			R_N^{\mathrm{anc}}\bm{g}_{Nj}
		\end{array}
	\right)
\end{eqnarray}
By using equation (\ref{qr-anchor-col-rep}), the objective function and constraint of equation (\ref{eig-problem}) can be replaced as follows.
\begin{equation}
	\begin{split}
		\min_{\bm{g}_{ij}} F(\bm{v}_j) &= {\bm{v}_j}^\top \bm{A}_{\tilde{X}} {\bm{v}_j} \\
		&= \sum^N_{i=1}\sum^N_{i'=1}\| \tilde{\bm{X}}_i^\mathrm{anc}\bm{g}_{ij} - \tilde{\bm{X}}_{i'}^\mathrm{anc}\bm{g}_{i'j} \|^2_2 \\
		&= \sum^N_{i=1}\sum^N_{i'=1}\| \bm{Q}_i^{\mathrm{anc}}(\bm{R}_i^{\mathrm{anc}}\bm{g}_{ij}) - \bm{Q}_{i'}^{\mathrm{anc}}(\bm{R}_{i'}^{\mathrm{anc}}\bm{g}_{i'j}) \|^2_2 \\
		&= {\bm{v}_j}'^\top \bm{A}_Q {\bm{v}_j}' \\
		\mathrm{s.t.} \; C(\bm{v}_j) &= {\bm{v}_j}^\top \bm{B}_{\tilde{X}} {\bm{v}_j} - 1 \\
		&= \sum^N_{i=1}\| \tilde{\bm{X}}_i^\mathrm{anc}\bm{g}_{ij} \|^2_2 - 1 \\
		&= \sum^N_{i=1}\| \bm{Q}_i^{\mathrm{anc}}(\bm{R}_i^{\mathrm{anc}}\bm{g}_{ij}) \|^2_2 - 1 \\
		&= {\bm{v}_j}'^\top \bm{B}_Q {\bm{v}_j}' - 1 = 0
	\end{split}
\end{equation}

Here, similar to the previously mentioned method, the method of Lagrange multipliers is applied. Since $\bm{A}_Q$ and $\bm{A}_{\tilde{X}}$, $\bm{B}_Q$ and $\bm{B}_{\tilde{X}}$ are matrices of the same size, and ${\bm{v}_j}$ and ${\bm{v}_j}'$ are vectors of the same size, this leads to an eigenvalue problem similar to equation (\ref{eig-l-3}).
\begin{equation}
	\label{eig-l-3-qr}
	\bm{A}_Q {\bm{v}_j}' = \lambda_j \bm{B}_Q {\bm{v}_j}' = \lambda_j {\bm{v}_j}' \quad ({\bm{v}_j}'^\top \bm{B}_Q {\bm{v}_j}' = {\bm{v}_j}'^\top {\bm{v}_j}' = 1)
\end{equation}
Furthermore, since $\bm{A}_Q$ is a symmetric matrix with a special matrix structure and can be decomposed into $\bm{W}_Q^{\top}\bm{W}_Q$, the eigenvalue problem in equation (\ref{eig-l-3-qr}) can be transformed as follows:
\begin{equation}
	\label{eq:eff-sol-1-qr}
	\begin{split}
		\bm{A}_Q {\bm{v}_j}' &= \lambda_j {\bm{v}_j}' \\
		-2(\bm{W}_{Q}^{\top}\bm{W}_{Q}) {\bm{v}_j}' + 2N{\bm{v}_j}' &= \lambda_j {\bm{v}_j}' \\
		\bm{W}_{Q}^{\top}\bm{W}_{Q} {\bm{v}_j}' &= \lambda_j' {\bm{v}_j}'
	\end{split}
\end{equation}
Where $\lambda'_j$ represents the following value:
\begin{eqnarray}
	\lambda'_j &=& - \frac{1}{2}(\lambda_j - 2N)
\end{eqnarray}
At this point, as equation (\ref{eq:eff-sol-1-qr}) is an eigenvalue problem concerning the symmetric matrix $\bm{W}_{Q}^{\top}\bm{W}_{Q}$, the eigenvalues $\lambda_j'$ and eigenvectors ${\bm{v}_j}'$ can be determined using SVD of $\bm{W}_{Q}$. Applying SVD to $\bm{W}_{Q}$ allows for the derivation of the solution (i.e., ${\lambda_j}'$, ${\bm{v}_j}'$) for the eigenvalue problem in equation (\ref{eq:eff-sol-1-qr}) and the solution (i.e., $\lambda_j$, $\bm{v}_j$) for the original generalized eigenvalue problem.
\begin{eqnarray}
	\bm{W}_Q &=& \bm{U} \bm{\Sigma} \bm{V}^\top = \bm{U} \left(
		\begin{array}{cccc}
			\sqrt{\lambda_1'} & & & O \\ 
			& \sqrt{\lambda_2'} & & \\ 
			& & \ddots & \\ 
			O & & & \sqrt{\lambda_N'}
		\end{array}
	\right) \left(
		\begin{array}{c}
			\bm{v}_1'^\top \\ 
			\bm{v}_2'^\top \\ 
			\vdots \\ 
			\bm{v}_N'^\top
		\end{array}
	\right) \\ 
	\lambda_j &=& -2\lambda_j'+2N \\ 
	\bm{v}_j &=& \left(
		\begin{array}{cccc}
			\bm{R}_1^\mathrm{anc} & & & O \\ 
			& \bm{R}_2^\mathrm{anc} & & \\ 
			& & \ddots & \\ 
			O & & & \bm{R}_N^\mathrm{anc}
		\end{array}
	\right)^{-1} \bm{v}_j'
\end{eqnarray}

\subsection{About Computational Complexity}
When solving the generalized eigenvalue problem to estimate the collaborative function, constructing the matrix to input into the solver of the generalized eigenvalue problem requires a computational complexity of $O(r(\tilde{m}N)^2)$, and solving the generalized eigenvalue problem itself takes a computational complexity of $O((\tilde{m}N)^3)$.

In contrast, when estimating the collaborative function through QR decomposition and SVD, constructing the matrix for applying SVD requires a computational complexity of $O(N(r\tilde{m})^2)$, and performing SVD takes a computational complexity of $O(r(\tilde{m}N)^2)$.

Solving the generalized eigenvalue problem directly is stable in terms of processing time with the increase of anchor data $r$, but it entails a significant increase in computational complexity as a cubic function with the increase in the dimension number of the intermediate representation $\tilde{m}$ and the number of institutions $N$. However, estimating the collaborative function using QR decomposition and SVD, while more affected by changes in the number of anchor data, keeps the increase in computational complexity due to the increase in the dimension number of the intermediate representation and the number of institutions compared to directly solving the generalized eigenvalue problem.

\section{Numerical experiments}
\label{sec:numerical_experiments}

In this section, we describe the setup for numerical experiments using real-world datasets. The analytical accuracy of DCA varies depending on the methods used for abstraction function, anchor data, and construction of the collaborative function. There are three research perspectives in this regard. However, as this study focuses on the construction of the collaborative function in DCA, we set the abstraction function as PCA and the anchor data as a random matrix following the normal distribution, and conduct experiments accordingly.

To compare the performance of the collaborative function optimization problem in DCA between the existing method \cite{dca-ori} and the proposed method, the following two types of experiments were conducted:
\begin{enumerate}
\item Compare the accuracy of various datasets using individual analysis, centralized analysis, existing method, and proposed method in DCA with a classification algorithm.
\item Compare the processing time for estimating the collaborative function from various datasets between the existing and proposed methods in DCA.
\end{enumerate}

\subsection{Datasets}
The experiments use three types of tabular datasets and one type of image dataset.

The first tabular dataset is the Mice Protein Expression Data Set \cite{dataset-1}, used for classifying the state of mice based on protein expression. The second is the QSAR biodegradation Data Set \cite{dataset-2}, for determining biodegradability from chemical structures. The third is the gene expression cancer RNA-Seq Data Set \cite{dataset-3}, used for classifying cancer based on gene expression information. With the gene dataset, using all 20,531 variables results in a prediction accuracy exceeding 99\%, making it difficult to observe performance differences. Therefore, experiments were conducted using only the first 200 dimensions of variables, which correspond to approximately 1\% of the total.

Additionally, the image dataset used is the CIFAR-10 dataset \cite{dataset-5}, for classifying images into ten categories, including vehicles and living organisms.

\subsection{Experimental Environment}
The experimental environment is as follows: \\
- Hardware: Apple M1, RAM 16GB \\
- OS: macOS \\
- Programming Language: Python \\
- Libraries Used: numpy, pandas, scikit-learn, scipy

\subsection{Experimental Setup}
This section explains the setup for the experiments. First, we describe the common settings across datasets, followed by specific settings for individual datasets.

\subsubsection{Common Settings for Datasets}
In each dataset's experiment, the data held by each institution is randomly distributed from the entire dataset. The values for the anchor data $\bm{X}^{\mathrm{anc}}$ are generated using random values following the standard normal distribution, and the method for constructing the intermediate representation is Principal Component Analysis (PCA).

\subsubsection{Experiments for Comparing Accuracy}
For measuring accuracy, five different data distribution patterns for each institution were prepared, and holdout validation was performed 10 times for each distribution pattern. The ratio of training data to test data in holdout validation is 50:50, and the distribution of training and test data in each of the 10 rounds is random for each distribution pattern.

Accuracy is used as the evaluation metric.

Two types of classification algorithms are used: Kernel SVC and Random Forest.

The number of anchor data $r$ is experimented in three patterns: 1$\times$, 3$\times$, and 9$\times$ of the dimension number of the original data for tabular data; and 1$\times$, 3$\times$, and 9$\times$ of the pixel number of the original data for image data.

The dimension number of the intermediate representation $\tilde{m}$ is experimented with three different numbers of dimensions where the cumulative contribution rate when PCA is applied to the original data distributed to institution 1 (i=1) is below 0.60, 0.75, and 0.90.

The comparison methods are as follows:
\begin{enumerate}
\item Individual analysis
\item Centralized analysis
\item DCA using minimal perturbation problem \cite{dca-ori}
\item DCA using generalized eigenvalue problem (Proposed Method)
\end{enumerate}
DCA using the generalized eigenvalue problem (Proposed Method) is experimented with two variants: one using the weighting method and the other without. However, weighting variables using the weighting method does not make sense in the case of Random Forest, so it is only used in experiments involving Kernel SVC.

\subsubsection{Experiments for Comparing Processing Time}
For experiments comparing processing times, only image data is used as the processing time for tabular data is extremely small and differences between methods are negligible.

For image data, the number of anchor data $r$ is experimented with in five patterns: 1$\times$, 3$\times$, 9$\times$, 27$\times$, and 81$\times$ the number of pixels in the original data.

The dimension number of the intermediate representation $\tilde{m}$ is set to the number of dimensions where the cumulative contribution rate is below 0.90 when PCA is applied to the original data distributed to institution 1 (i=1), to compare processing times under larger scale conditions.

For each measurement point, experiments are conducted once for each of the five data distribution patterns to each institution, and the average values are calculated.

The comparison methods are as follows:
\begin{enumerate}
\item Minimal Perturbation Problem (SVD) \cite{dca-ori}
\item Minimal Perturbation Problem (Randomized SVD) \cite{dca-ori}
\item Generalized Eigenvalue Problem (Proposed Method)
\item QR Decomposition + SVD (Proposed Method)
\item QR Decomposition + Randomized SVD (Proposed Method)
\end{enumerate}

\subsection{Mice Protein Expression Dataset}
The total number of data across all institutions is $n=1000$, with $N=20$ institutions, and 50 data points per institution.
The dimension number of the original data is $m=77$.

\subsection{QSAR (QSAR Biodegradation) Dataset}
The total number of data across all institutions is $n=1000$, with $N=20$ institutions, and 50 data points per institution.
The dimension number of the original data is $m=41$.

\subsection{Gene Expression (Gene Expression Cancer RNA-Seq) Dataset}
The total number of data across all institutions is $n=800$, with $N=16$ institutions, and 50 data points per institution.
The dimension number of the original data is $m=200$.

\subsection{Vehicle and Biological Images (CIFAR-10) Dataset}
For accuracy comparison, the total number of data across all institutions is $n=2000$, with $N=20$ institutions, 100 data points per institution, and the dimension number of the original data is $m=3072$, with 1024 pixels.
For processing time comparison, to test under larger scales, the total number of data across all institutions and the number of institutions are both increased to $n=8000$ and $N=80$ respectively.

\section{Results}
\label{sec:results}

In this section, we present the results and discussion of the numerical experiments.

\subsection{Experiments for Comparing Accuracy}
In the experiments comparing accuracy, the change in accuracy when varying only one of the parameters the number of institutions, the dimension number of the intermediate representation, and the number of anchor data is shown in separate graphs. In each graph, when varying one parameter, the other two parameters are set to the maximum number of institutions $N$, the dimension number of the intermediate representation $\tilde{m}$ set to the dimension number below a cumulative contribution rate of 0.90 when PCA is applied to the original data distributed to institution 1 (i=1), and the number of anchor data $r$ set to 9$\times$ the dimension number of the original data for tabular data and 9$\times$ the number of pixels for image data.

\subsubsection{Mice Protein Expression Dataset}
The average accuracy for individual analysis was 0.456 for Kernel SVC and 0.406 for Random Forest.

\begin{figure}[htbp]
	\begin{minipage}{0.33\hsize}
		\begin{center}
      \includegraphics[scale=0.33]{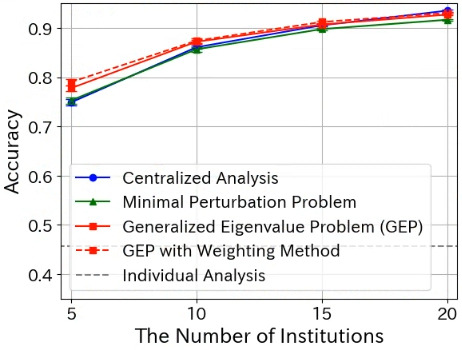}
		\end{center}
		\subcaption{Institutions Variation}
	\end{minipage}
	\begin{minipage}{0.33\hsize}
		\begin{center}
      \includegraphics[scale=0.33]{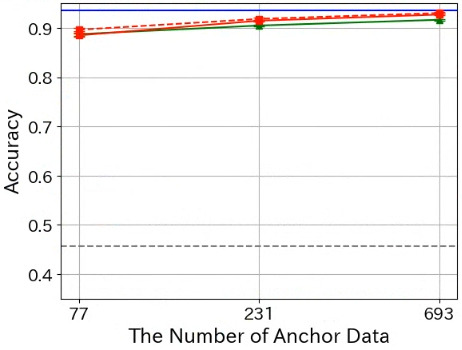}
		\end{center}
		\subcaption{Anchor Data Variation}
	\end{minipage}
  \begin{minipage}{0.33\hsize}
		\begin{center}
      \includegraphics[scale=0.33]{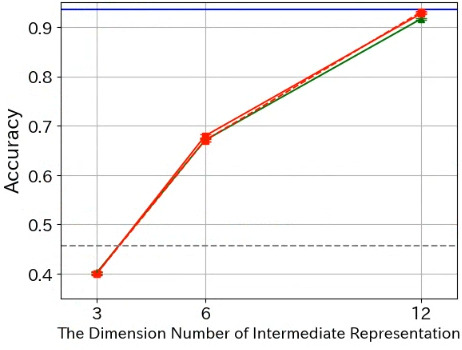}
		\end{center}
		\subcaption{Dimension Number Variation}
	\end{minipage}

	\caption{Accuracy for the Mice Protein Expression Dataset (Kernel SVC)}
	\label{figure:acc-mice-1}
\end{figure}

\begin{figure}[htbp]
	\begin{minipage}{0.33\hsize}
		\begin{center}
      \includegraphics[scale=0.33]{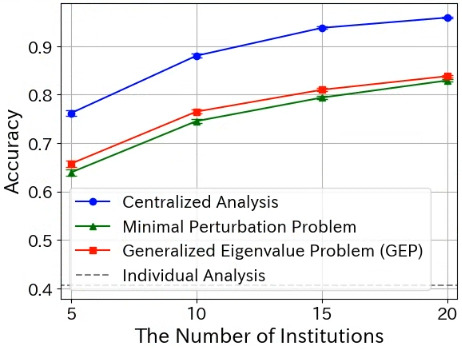}
		\end{center}
		\subcaption{Institutions Variation}
	\end{minipage}
	\begin{minipage}{0.33\hsize}
		\begin{center}
      \includegraphics[scale=0.33]{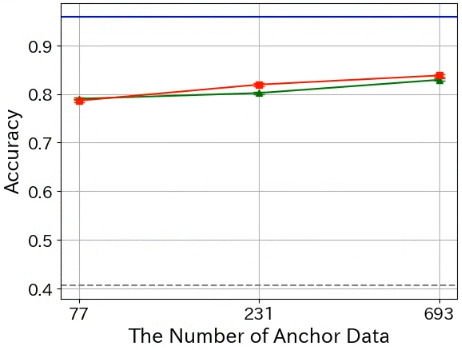}
		\end{center}
		\subcaption{Anchor Data Variation}
	\end{minipage}
  \begin{minipage}{0.33\hsize}
		\begin{center}
      \includegraphics[scale=0.33]{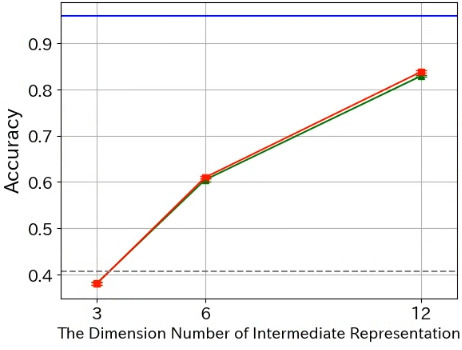}
		\end{center}
		\subcaption{Dimension Number Variation}
	\end{minipage}

	\caption{Accuracy for the Mice Protein Expression Dataset (Random Forest)}
	\label{figure:acc-mice-2}
\end{figure}

First, the graph comparing accuracy with Kernel SVC as the classification algorithm is shown in Figure \ref{figure:acc-mice-1}. The vertical axis represents accuracy.
Similarly, the graph comparing accuracy with Random Forest as the classification algorithm is shown in Figure \ref{figure:acc-mice-2}.

From Figures \ref{figure:acc-mice-1} and \ref{figure:acc-mice-2}, it can be observed that when the dimension number of the intermediate representation is above a certain threshold, DCA methods demonstrate superior accuracy compared to individual analysis, similar to centralized analysis. The accuracy of DCA methods improves as the number of institutions, the number of anchor data, and the dimension number of the intermediate representation increase. Furthermore, when comparing DCA methods, the proposed method using the generalized eigenvalue problem showed the best performance for this dataset.

\subsubsection{QSAR (QSAR Biodegradation) Dataset}
The average accuracy for individual analysis was 0.716 for Kernel SVC and 0.754 for Random Forest.

\begin{figure}[htbp]
	\begin{minipage}{0.33\hsize}
		\begin{center}
      \includegraphics[scale=0.33]{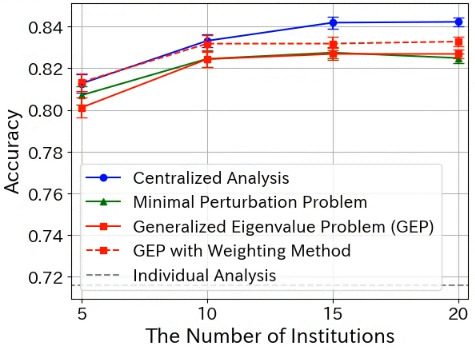}
		\end{center}
		\subcaption{Institutions Variation}
	\end{minipage}
	\begin{minipage}{0.33\hsize}
		\begin{center}
      \includegraphics[scale=0.33]{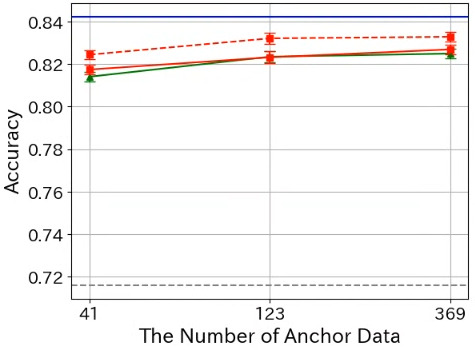}
		\end{center}
		\subcaption{Anchor Data Variation}
	\end{minipage}
  \begin{minipage}{0.33\hsize}
		\begin{center}
      \includegraphics[scale=0.33]{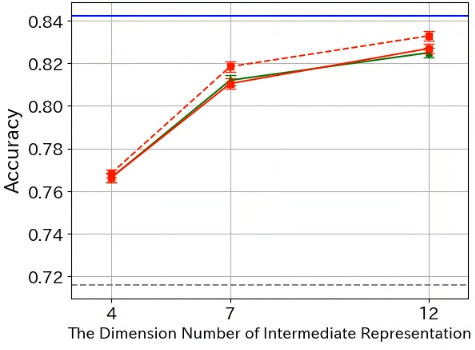}
		\end{center}
		\subcaption{Dimension Number Variation}
	\end{minipage}

	\caption{Accuracy for the QSAR Dataset (Kernel SVC)}
	\label{figure:acc-qsar-1}
\end{figure}

First, the graph comparing accuracy with Kernel SVC as the classification algorithm is shown in Figure \ref{figure:acc-qsar-1}. The vertical axis represents accuracy.

\begin{figure}[htbp]
  \begin{minipage}{0.33\hsize}
		\begin{center}
      \includegraphics[scale=0.33]{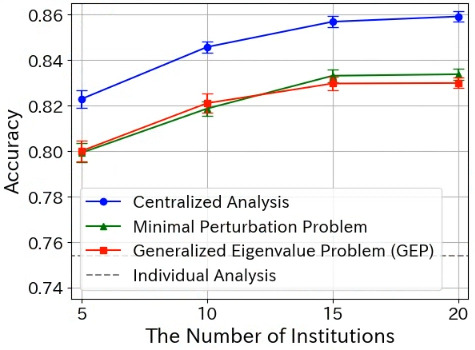}
		\end{center}
		\subcaption{Institutions Variation}
	\end{minipage}
	\begin{minipage}{0.33\hsize}
		\begin{center}
      \includegraphics[scale=0.33]{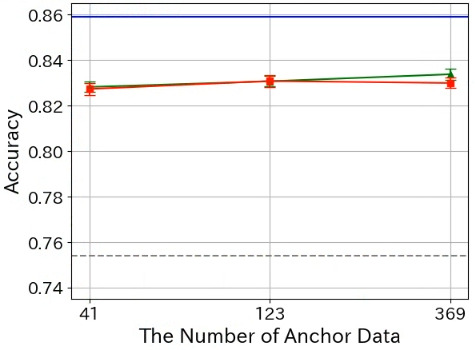}
		\end{center}
		\subcaption{Anchor Data Variation}
	\end{minipage}
  \begin{minipage}{0.33\hsize}
		\begin{center}
      \includegraphics[scale=0.33]{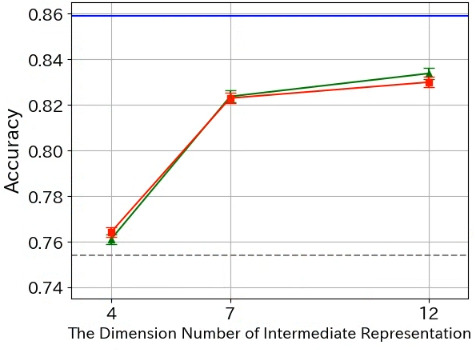}
		\end{center}
		\subcaption{Dimension Number Variation}
	\end{minipage}

	\caption{Accuracy for the QSAR Dataset (Random Forest)}
	\label{figure:acc-qsar-2}
\end{figure}

Similarly, the graph comparing accuracy with Random Forest as the classification algorithm is shown in Figure \ref{figure:acc-qsar-2}.

From Figures \ref{figure:acc-qsar-1} and \ref{figure:acc-qsar-2}, it is observed that DCA methods demonstrate superior accuracy compared to individual analysis, similar to centralized analysis.  The accuracy of DCA methods improves as the number of institutions, the number of anchor data, and the dimension number of the intermediate representation increase. Furthermore, when comparing DCA methods for this dataset, the proposed method using the generalized eigenvalue problem with the weighting method showed the best performance when Kernel SVC is used as the classification algorithm, confirming the effectiveness of the weighting method. However, when Random Forest is used as the classification algorithm, the performance differences between methods are minimal. The inability to apply the weighting method and attenuate the features of poor quality collaborative representations may explain why the proposed method using the generalized eigenvalue problem was not as effective with Random Forest.

\subsubsection{Gene Expression (Gene Expression Cancer RNA-Seq) Dataset}
The average accuracy for individual analysis was 0.961 for Kernel SVC and 0.813 for Random Forest.

\begin{figure}[htbp]
  \begin{minipage}{0.33\hsize}
		\begin{center}
      \includegraphics[scale=0.29]{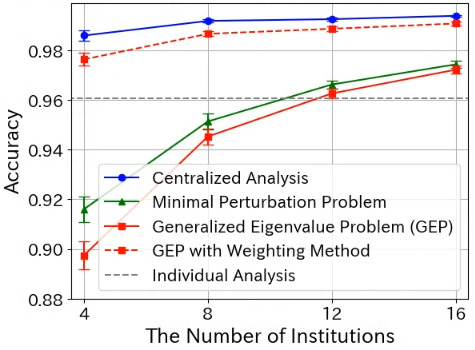}
		\end{center}
		\subcaption{Institutions Variation}
	\end{minipage}
	\begin{minipage}{0.33\hsize}
		\begin{center}
      \includegraphics[scale=0.29]{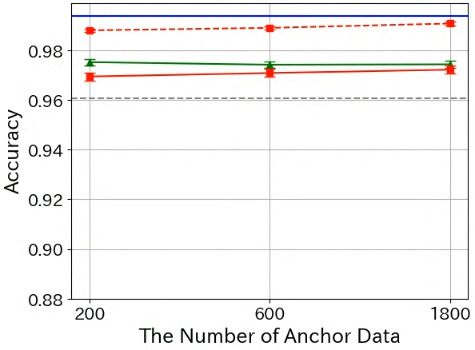}
		\end{center}
		\subcaption{Anchor Data Variation}
	\end{minipage}
  \begin{minipage}{0.33\hsize}
		\begin{center}
      \includegraphics[scale=0.29]{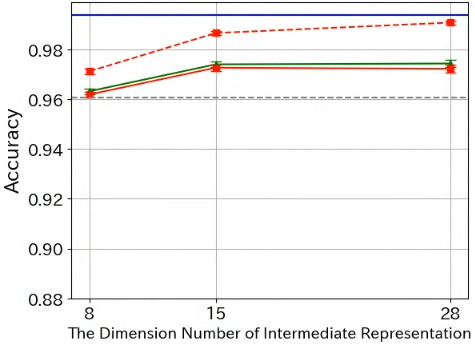}
		\end{center}
		\subcaption{Dimension Number Variation}
	\end{minipage}

	\caption{Accuracy for the Gene Expression Dataset (Kernel SVC)}
	\label{figure:acc-gene-1}
\end{figure}

First, the graph comparing accuracy with Kernel SVC as the classification algorithm is shown in Figure \ref{figure:acc-gene-1}. The vertical axis represents accuracy.

\begin{figure}[htbp]
  \begin{minipage}{0.33\hsize}
		\begin{center}
      \includegraphics[scale=0.32]{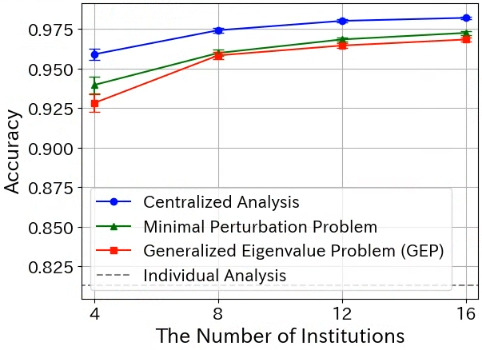}
		\end{center}
		\subcaption{Institutions Variation}
	\end{minipage}
	\begin{minipage}{0.33\hsize}
		\begin{center}
      \includegraphics[scale=0.32]{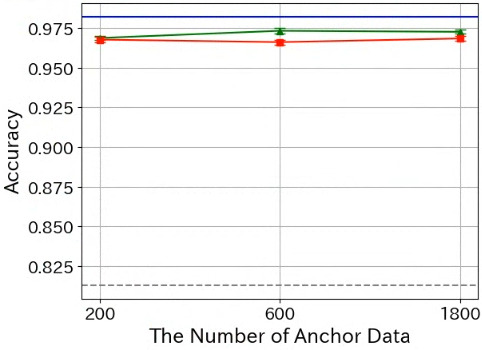}
		\end{center}
		\subcaption{Anchor Data Variation}
	\end{minipage}
  \begin{minipage}{0.33\hsize}
		\begin{center}
      \includegraphics[scale=0.32]{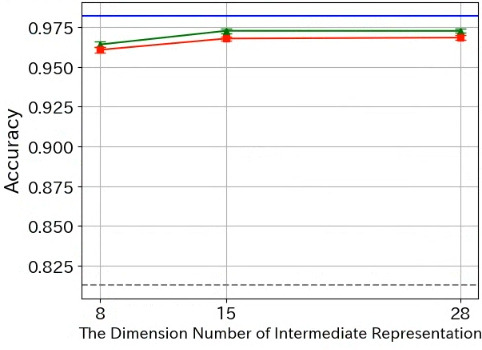}
		\end{center}
		\subcaption{Dimension Number Variation}
	\end{minipage}

	\caption{Accuracy for the Gene Expression Dataset (Random Forest)}
	\label{figure:acc-gene-2}
\end{figure}

Similarly, the graph comparing accuracy with Random Forest as the classification algorithm is shown in Figure \ref{figure:acc-gene-2}.

From Figures \ref{figure:acc-gene-1} and \ref{figure:acc-gene-2}, it is observed that DCA methods demonstrate superior accuracy compared to individual analysis, similar to centralized analysis, when the number of institutions is above a certain threshold. The accuracy of DCA methods improves as the number of institutions, the number of anchor data, and the dimension number of the intermediate representation increase. Furthermore, when comparing DCA methods for this dataset, the proposed method using the generalized eigenvalue problem with the weighting method showed the best performance when Kernel SVC is used as the classification algorithm. However, as with the previous datasets, since the weighting method could not attenuate the features of poor quality collaborative representations, the proposed method using the generalized eigenvalue problem was not as effective when Random Forest was used as the classification algorithm.


\subsubsection{Vehicles and Living Organisms Images (CIFAR-10) Dataset}
The average accuracy for individual analysis was 0.196 for Kernel SVC and 0.178 for Random Forest.

\begin{figure}[htbp]
  \begin{minipage}{0.33\hsize}
		\begin{center}
      \includegraphics[scale=0.32]{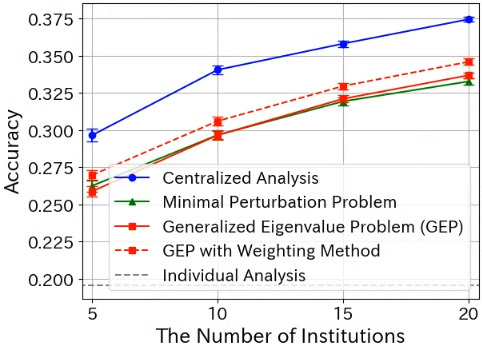}
		\end{center}
		\subcaption{Institutions Variation}
	\end{minipage}
	\begin{minipage}{0.33\hsize}
		\begin{center}
      \includegraphics[scale=0.32]{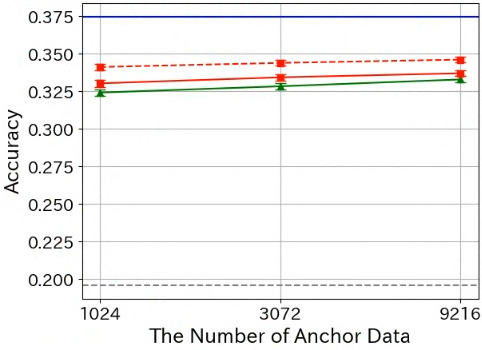}
		\end{center}
		\subcaption{Anchor Data Variation}
	\end{minipage}
  \begin{minipage}{0.33\hsize}
		\begin{center}
      \includegraphics[scale=0.32]{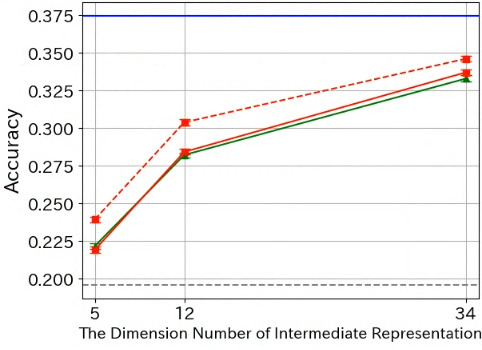}
		\end{center}
		\subcaption{Dimension Number Variation}
	\end{minipage}

	\caption{Accuracy for the Vehicles and Living Organisms Images Dataset (Kernel SVC)}
	\label{figure:acc-cifar-1}
\end{figure}

\begin{figure}[htbp]
	\begin{minipage}{0.33\hsize}
		\begin{center}
      \includegraphics[scale=0.32]{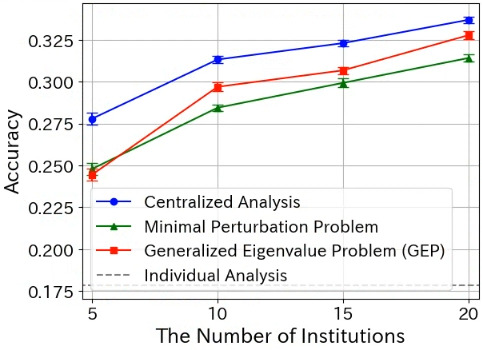}
		\end{center}
		\subcaption{Institutions Variation}
	\end{minipage}
	\begin{minipage}{0.33\hsize}
		\begin{center}
      \includegraphics[scale=0.32]{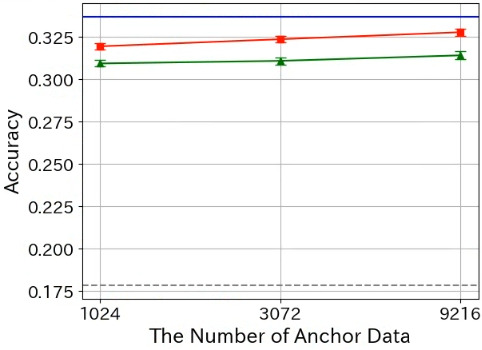}
		\end{center}
		\subcaption{Anchor Data Variation}
	\end{minipage}
  \begin{minipage}{0.33\hsize}
		\begin{center}
      \includegraphics[scale=0.32]{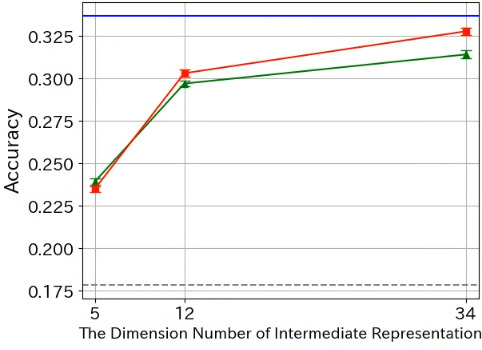}
		\end{center}
		\subcaption{Dimension Number Variation}
	\end{minipage}

	\caption{Accuracy for the Vehicles and Living Organisms Images Dataset (Random Forest)}
	\label{figure:acc-cifar-2}
\end{figure}

First, the graph comparing accuracy with Kernel SVC as the classification algorithm is shown in Figure \ref{figure:acc-cifar-1}. The vertical axis represents accuracy.
Similarly, the graph comparing accuracy with Random Forest as the classification algorithm is shown in Figure \ref{figure:acc-cifar-2}.

From Figures \ref{figure:acc-cifar-1} and \ref{figure:acc-cifar-2}, it is observed that DCA methods demonstrate superior accuracy compared to individual analysis, similar to centralized analysis. The accuracy of DCA methods improves as the number of institutions and the dimension number of the intermediate representation increase. Furthermore, when comparing DCA methods for this dataset, the proposed method using the generalized eigenvalue problem showed the best performance regardless of whether Kernel SVC or Random Forest was used as the classification algorithm.

\subsection{Experiments for Comparing Processing Time} \label{result-speed}
In the experiments comparing processing time, the change in processing time when varying only one of the parameters either the number of institutions or the number of anchor data is shown in separate graphs. In each graph, when varying one parameter, the other parameter is fixed: the number of institutions $N$ is set to the 20 of institutions, and the number of anchor data $r$ is set to 9$\times$ the number of pixels for image data.

\subsubsection{Vehicles and Living Organisms Images (CIFAR-10) Dataset}
\begin{figure}[htbp]
	\begin{minipage}{0.5\hsize}
		\begin{center}
      \includegraphics[scale=0.28]{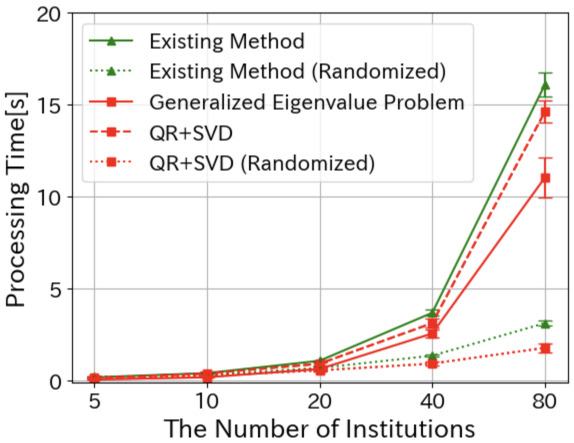}
		\end{center}
		\subcaption{Institutions Variation}
	\end{minipage}
	\begin{minipage}{0.5\hsize}
		\begin{center}
      \includegraphics[scale=0.28]{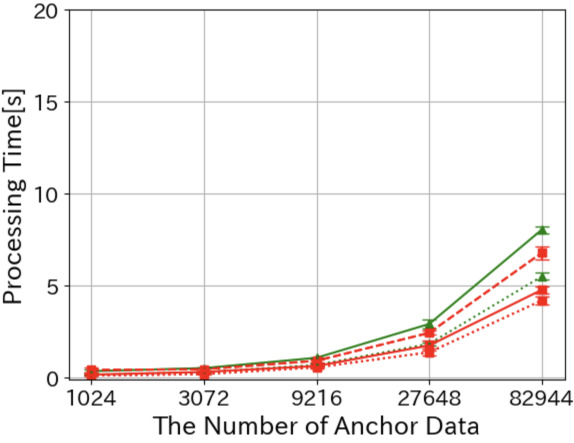}
		\end{center}
		\subcaption{Anchor Data Variation}
	\end{minipage}

	\caption{Processing Time for the Vehicles and Living Organisms Images Dataset (The Dimension Number of Intermediate Representation:\; 34)}
	\label{figure:time-cifar-4}
\end{figure}

The graph comparing processing times with the dimension number of the intermediate representation set to 34 (the number of dimensions below a cumulative contribution rate of 0.90 when PCA is applied to the original data distributed to institution 1) is shown in Figure \ref{figure:time-cifar-4}. The vertical axis represents processing time.

From Figure \ref{figure:time-cifar-4}, it is evident that for the Vehicle and Living Organisms Images dataset with original pixel count of 1024 and an intermediate representation of 34 dimensions, differences between methods are observable, unlike with the tabular datasets. Both the existing minimal perturbation problem and the methods using QR decomposition + SVD are faster when using randomized algorithms. Furthermore, the generalized eigenvalue problem has a smaller impact on processing speed due to the number of anchor data in terms of computational complexity, making it particularly fast and comparable with randomized algorithm when there is a large amount of anchor data. Therefore, the proposed method can efficiently determine the collaborative function by selecting solutions based on the scale of the problem, such as the number of anchor data and institutions.

\subsection{Overall Experimental Results}
From the overall results comparing accuracy, it has been confirmed that the exact solutions for the collaborative function obtained by solving the proposed generalized eigenvalue problem are particularly high-performing compared to the existing methods using the minimal perturbation problem. However, it was observed in Figures \ref{figure:acc-gene-1} and \ref{figure:acc-gene-2} that the performance of the method using the generalized eigenvalue problem without the weighting method was low. Therefore, there are challenges in methods that attenuate the impact of poor quality collaborative representations in formulations using the generalized eigenvalue problem. In the case of Kernel SVC, which has penalty terms for weights, applying the weighting method to the solution using the generalized eigenvalue problem resulted in better accuracy.

Furthermore, from the comparison of processing times, it has been shown that as the number of institutions and the number of anchor data increase, the processing time to determine the collaborative function also increases. In the experiments with the larger image dataset, as seen in Figure \ref{figure:time-cifar-4}, it was confirmed that methods using randomized algorithms are faster, and the proposed method can efficiently determine the collaborative function by selecting solutions based on the scale of the problem, such as the number of anchor data and institutions.

\section{Conclusions}
\label{sec:conclusions}

In this section, we present the conclusions and future challenges of this study.

In this research, we formulated a new decomposition of the matrix $\bm{G}_i$, which corresponds to the collaborative function, for the collaborative function optimization problem, and proposed a solution method using the generalized eigenvalue problem. The proposed method using the generalized eigenvalue problem includes an effective construction method for the collaborative function with the weighting method and a solution method reducing to SVD, which is efficient when the number of anchor data is small.

In experiments using real-world datasets, we confirmed that the collaborative functions obtained by solving the proposed generalized eigenvalue problem are particularly high-performing across all datasets. Especially in models with penalty terms, like Kernel SVC, the weighting method proved to be particularly effective. Additionally, we confirmed that the proposed method using the generalized eigenvalue problem has processing speeds comparable to existing methods. Particularly in situations involving large datasets, like image data, the proposed method efficiently determines the collaborative function by choosing whether to solve the generalized eigenvalue problem directly or reduce it to SVD based on factors such as the number of anchor data.

A key point of the new formulation is that it divides the problem into columns and applies norm constraints to the column vectors. The main contributions of this study are as follows:
\begin{enumerate}
	\item Simplification of the solution process for the collaborative function optimization problem.
	\item In numerical experiments, the proposed formulation and solution method for the collaborative function achieved higher performance than existing methods.
	\item Efficiency in determining the collaborative function by selecting solutions according to the situation.
\end{enumerate}

Utilizing nonlinear abstraction functions and collaborative functions for further accuracy improvement is a significant future challenge. Additionally, speeding up processing times even when the number of institutions and dimension number of the intermediate representation are large is also necessary for broader application of DCA.

\bibliographystyle{elsarticle-num}




\end{document}